\pdfoutput=1

\documentclass[11pt]{article}

\usepackage{ACL2023}

\usepackage{times}
\usepackage{latexsym}

\usepackage[T1]{fontenc}

\usepackage[utf8]{inputenc}

\usepackage{microtype}

\usepackage{inconsolata}

\usepackage{multirow}
\usepackage{makecell}

\usepackage{times}
\usepackage{latexsym}
\usepackage{graphicx}
\usepackage{subfigure}
\usepackage{float}
\graphicspath{{images}}
\usepackage{stfloats}
\usepackage{booktabs}
\usepackage{indentfirst}
\usepackage{amsmath}
\usepackage{amssymb}

\usepackage{enumitem}

\usepackage{colortbl}
\usepackage{arydshln}

\usepackage[linesnumbered,ruled,vlined]{algorithm2e}

\SetCommentSty{mycommfont}

\SetKwInput{KwInput}{Input}                
\SetKwInput{KwOutput}{Output}              

\usepackage{CJKutf8}

\usepackage{listings}
\usepackage{xcolor}

\definecolor{codegreen}{rgb}{0,0.6,0}
\definecolor{codegray}{rgb}{0.5,0.5,0.5}
\definecolor{codepurple}{rgb}{0.58,0,0.82}
\definecolor{backcolour}{rgb}{0.95,0.95,0.92}

\lstdefinestyle{mystyle}{
    backgroundcolor=\color{backcolour},   
    commentstyle=\color{codegreen},
    keywordstyle=\color{magenta},
    numberstyle=\tiny\color{codegray},
    stringstyle=\color{codepurple},
    basicstyle=\ttfamily\small,
    breakatwhitespace=false,         
    breaklines=true,                 
    captionpos=b,                    
    keepspaces=true,                 
    numbers=left,                    
    numbersep=5pt,                  
    showspaces=false,                
    showstringspaces=false,
    showtabs=false,                  
    tabsize=2
}

\title{DRUM: Learning \underline{D}emonstration \underline{R}etriever for Large M\underline{U}lti-modal \underline{M}odels }

\author{
Ellen Yi-Ge$^1$ \ \ 
Jiechao Gao$^2$ \ \
Wei Han$^3$ \ \
\textbf{Wei Zhu}$^4$\thanks{\ \ Corresponding author. For any inquiries, please contact: michaelwzhu91@gmail.com; } \\ 
\textsuperscript{\rm 1} Carnegie Mellon University, PA, United States  \\
\textsuperscript{\rm 2} University of Virginia, VA, United States  \\
\textsuperscript{\rm 3} Independent Researcher, TX, United States  \\
\textsuperscript{\rm 4} University of Hong Kong, HK, China \\
}

\begin{document}
\maketitle
\begin{abstract}

Recently, large language models (LLMs) have demonstrated impressive capabilities in dealing with new tasks with the help of in-context learning (ICL). In the study of Large Vision-Language Models (LVLMs), when implementing ICL, researchers usually adopts the naive strategies like fixed demonstrations across different samples, or selecting demonstrations directly via a visual-language embedding model. These methods does not guarantee the configured demonstrations fit the need of the LVLMs. To address this issue, we now propose a novel framework, \underline{d}emonstration \underline{r}etriever for large m\underline{u}lti-modal \underline{m}odel (DRUM), which fine-tunes the visual-language embedding model to better meet the LVLM's needs. First, we discuss the retrieval strategies for a visual-language task, assuming an embedding model is given. And we propose to concate the image and text embeddings to enhance the retrieval performance. Second, we propose to re-rank the demonstrations retrieved by the embedding model via the LVLM's feedbacks, and calculate a list-wise ranking loss for training the embedding model. Third, we propose an iterative demonstration mining strategy to improve the training of the embedding model. Through extensive experiments on 3 types of visual-language tasks, 7 benchmark datasets, our DRUM framework is proven to be effective in boosting the LVLM's in-context learning performance via retrieving more proper demonstrations.  

\end{abstract}

\begin{CJK*}{UTF8}{gbsn}

\section{Introduction}

In-context learning (ICL) is a simple yet important learning paradigm that given a few input-output pairs (demonstrations), a model can learn to conduct predictions on a new task it never sees before. ICL is a type of emergent capability observed in large-scale pre-trained models \cite{wei2022emergent}. It is first observed by GPT-3 \cite{brown2020language}, and draws the attention of the whole community of artificial intelligence. And a large branch of literature have shown that large language models (LLMs) have impressive ICL capabilities across a wide range natural language processing (NLP) tasks. ICL is essential for applications, since it can quickly adapt the large pretrained models to a novel task, or a task with personalized needs, with only a few demonstrations. No fine-tuning is needed and the model need not to be deployed again.

Recently, large vision-language models (LVLMs) are being rapidly developed, and its ICL capabilities are also being investigated \cite{alayrac2022flamingo}. The LVLMs like Flamingo \cite{alayrac2022flamingo} and Qwen-VL \cite{bai2023qwen} have demonstrated impressive ICL capabilities on the visual question answering (VQA), few-shot image classification (ImageCLS), and image captioning (ImageCAP) tasks. However, when implementing ICL for LVLMs, researchers usually adopts the naive strategies like fixed demonstrations or demonstrations ranked by a pre-trained visiion-language embedding model. These strategies are sub-optimal, since they do not incorporate the LVLMs' feedbacks on how these demonstrations help them to improve the responses.

\begin{figure*}[t]
\centering
\includegraphics[width=0.76\textwidth]{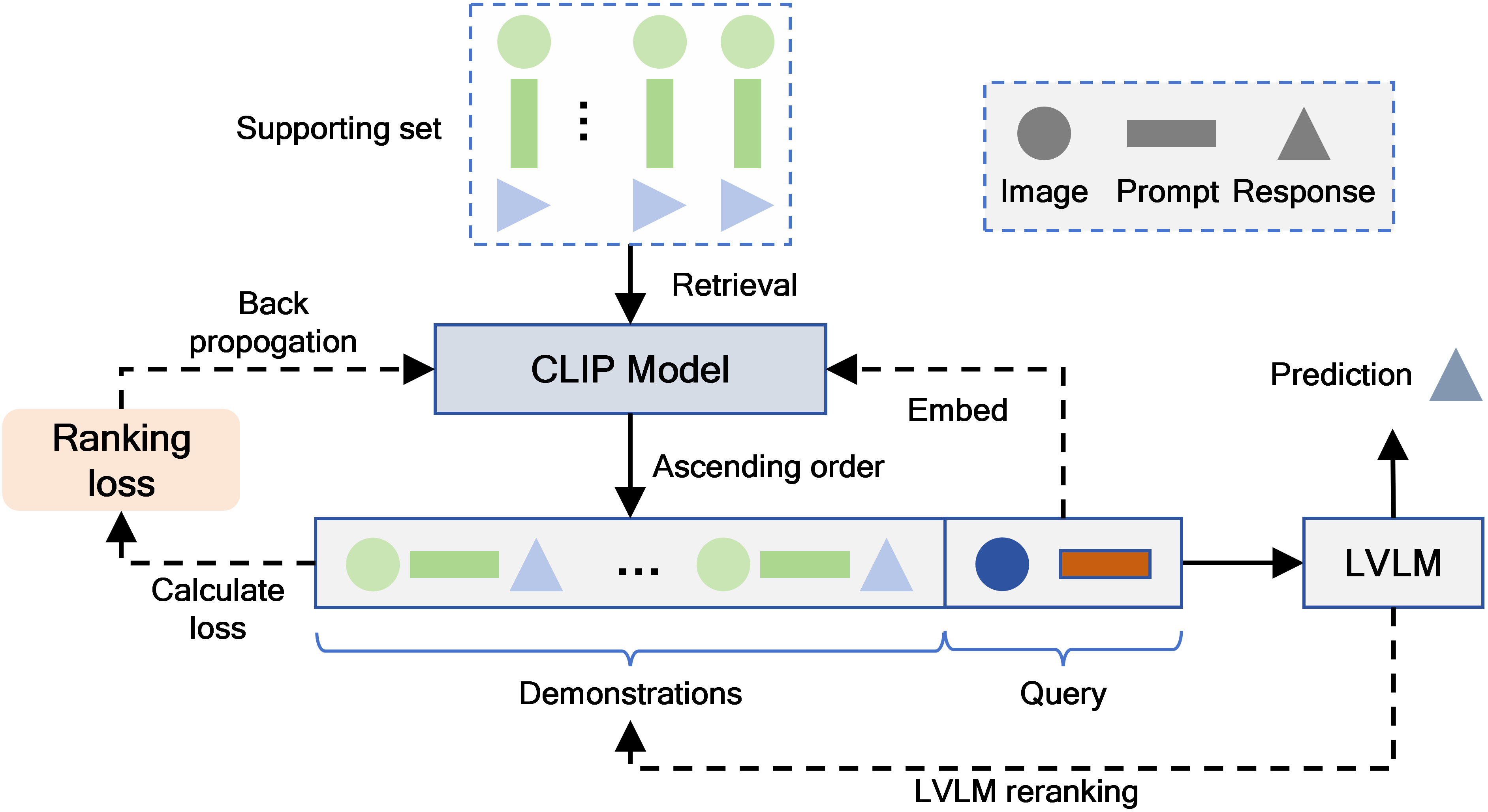} 
\caption{The schematic representation of our DRUM framework. Circles, rectangles, and triangles respectively represent the images, prompts, and responses in the triplet. }
\label{fig:framework}
\end{figure*}

To address the above issue, we now present a novel framework, \underline{d}emonstration \underline{r}etriever for large m\underline{u}lti-modal \underline{m}odel (DRUM). DRUM is targeted at fine-tuning a pre-trained visual-language embedding model so that it learns to retrieve better demonstrations to meet the LVLM's needs when conduct inference. First, assuming the embedding model is given, DRUM discusses the retrieval strategy for any visual-language tasks. And it proposes to retrieve demonstrations based on the joint embedding of input image, prompt and draft response. Second, DRUM asks the inference LVLM to re-rank the embedding model's retrieved demonstrations via the LVLM feedback. In this work, the LVLM feedback on a demonstration is defined as the conditional log-likelihood of the target response when the demonstration is added to the prompt. With the LVLM's reranking results, a list-wise ranking loss can be calculated and used as the optimization objective for the embedding model. Third, we propose an iterative demonstration mining strategy which updates the demonstration candidates iteratively, thus improving the training of the embedding model by providing high-quality ranking signals.

We have conducted extensive experiments on 3 types of visual-language tasks, VQA, ImageCLS and ImageCAP, and totally 7 benchmark datasets. The experimental results demonstrate that our DRUM framework is effective in boosting the LVLM's ICL performance. In addition, for commercial LVLMs like GPT-4o, the embedding model fine-tuned by DRUM can also be transferred to them, help them to retrieve better demonstrations. 

Our contributions are as follows: 
\begin{itemize}
\item We propose a novel framework, DRUM, to enhance the ICL capabilities of the LVLMs.

\item Extensive experiments have proven that DRUM is effective in boosting the LVLMs' ICL performance on a wide range of vision-language tasks. 

\end{itemize}

\section{Related Work}

\noindent\textbf{In-Context Learning in NLP.} The artificial intelligence community has witnessed significant advancements in the realm of large language models (LLMs) in recent years. As these models and their training corpora expand in scale, LLMs have demonstrated emergent capabilities, such as reasoning, mathematical proficiency, and the ability to follow prompts \cite{wei2022emergent}. GPT-3 \cite{brown2020language} was the pioneer in revealing that sufficiently large models can learn to execute new tasks with minimal guidance, a phenomenon termed in-context learning (ICL). Subsequent studies have corroborated the impressive performance of LLMs across various tasks through ICL \cite{mosbach2023few}. The crux of ICL lies in the construction of high-quality in-context demonstration sequences \cite{Li2023UnifiedDR}. However, the bulk of these explorations have concentrated on pure natural language processing tasks and text-centric foundation models, highlighting the necessity to extend this research to encompass other domains.

The research works on in-context learning focus primarily on demonstration sequences. A series of techniques have been investigated, including: (a) utilizing similarity scores to retrieve more relevant in-context examples \cite{Li2023UnifiedDR},  (b) employing machine-generated demonstrations \cite{li2023human}. The literature has seen a series of studies that reveals certain properties of LLMs when applied to in-context learning. \citet{pan2023context} proposed a decomposition of ICL into the task recognition effect and the task learning effect, and quantified these capabilities of models with varying numbers of shots and scales. Additionally, \citet{lyu2022z} records the "copying effect" phenomenon in LLMs, which is also a type of shortcut inference. Our work complements this line of research by fine-tuning the vision-language embedding model to learn how to retrieve appropriate demonstrations.

\noindent \textbf{LVLM and ICL} \quad Inspired by the triumphs of LLMs in natural language processing, the vision-language domain has seen the rise of analogous large vision-language models (LVLMs) \cite{du2022survey}. Among these, models such as BLIP2 \cite{li2023blip}, MiniGPT-4 \cite{zhu2023minigpt}, and LLAVA \cite{liu2024visual} are pretrained by aligning image and text data through the use of adapters \cite{houlsby2019parameter} to reduce training overhead. While there are several VLMs available, it is worth noting that some of the models are unsuitable for in-context learning, as this capability demands that the LVLM handle inputs that interweave images and text content \cite{alayrac2022flamingo}. Presently, there is scant research on multimodal ICL or ICL for LVLMs, with only a few studies focusing on rudimentary strategies. \citet{yang2024exploring} examines the impact of ICL on the LVLM's performance in image captioning tasks. \citet{li2024configure} analyzes the effects of ICL for LVLMs and proposes various strategies for demonstration retrieval using a pre-trained vision-language embedding model, such as CLIP \cite{radford2021learning}. Our work complements this line of research by proposing a novel framework for ICL of the LVLMs.

\section{DRUM}

\noindent \textbf{Formulation} \quad Given a well pre-trained Large Vision-Language Model (LVLM) (denoted as $\mathcal{M}$) e.g., Flamingo \cite{alayrac2022flamingo}, one can use it directly to solve a VL task like VQA with in-context learning, and no fine-tuning is required. To achieve this, we need to prepare a multi-modal in-context sequence 
\begin{equation}
\mathcal{S} = \{ z_{1}, ..., z_{n} \},
\end{equation}
where $\mathcal{S}$ consists of $n$-shot $z_{i} = $ ($\text{image}_{i}$, $\text{prompt}_{i}$, $\text{response}_{i}$) tuples. Then we concatenate $\mathcal{S}$ to the left of the test sample $x_{test} = $ ($\text{image}_{test}$, $\text{prompt}_{test}$), and feed into the LVLM for generating the corresponding response: 
\begin{equation}
\text{response}_{test} = \{ \hat{a}_{1}, ..., \hat{a}_{T_{A}}  \},
\end{equation}
where the $t$-th ($t \leq T_{A} $) token $\hat{a}_{t}$ is sampled from the probability distribution $\mathbf{P}(\cdot)$ over the vocabulary calculated by the LVLM $\mathcal{M}$: 
\begin{equation}
\mathbf{P}( \hat{a}_{t} | \mathcal{S}, x_{test}, \hat{a}_{1:t-1} ).
\end{equation}

\subsection{Strategies to retrieve demonstrations}
\label{subsec:demon_retrievals}

Note that each component (image, prompt, and response) of a sample can be used to construct the vector database and build an index, and we can respectively use them to retrieve $n$ examples from the supporting set $\mathcal{D} = \{ z_1, z_2, ..., z_N\}$ ($n < N$) as the demonstrations for $n$-shot setting. Then, we can use these triplets to form the in-context sequence $S$. Now, we summarize a series of specific retrieval strategies mentioned in the literature \cite{li2024configure} and new ones proposed in our work .

\noindent \textbf{Random sampling (RS)} This strategy simply obeys the uniform distribution to randomly sample n-shot triplets from $\mathcal{D}$ to form the in-context sequence $S$.

\noindent \textbf{Retrieving via similar image (SI)} This method retrieve $n$ images from $\mathcal{D}$ which are most similar to the querying image and then use the corresponding triplets of these retrieved images as the demonstrations. For example, given the test sample $x_{test} = $ ($\text{image}_{test}$, $\text{prompt}_{test}$), suppose the $i$-th image $\text{image}_{i}$ is similar to $\text{image}_{test}$, then the whole $i$-th triplet $z_{i} = $ ($\text{image}_{i}$, $\text{prompt}_{i}$, $\text{response}_{i}$) will be used as one demonstration. Here we assume we have access to an high-quality image embedding model at hand, which can transform each image to a separate vector in the semantic space in which the similarity between two vectors reflect their similarity in contents.

\noindent \textbf{Retrieving via similar texts (ST).} Besides retrieving via images, we can also retrieve $n$ triplets which contain the most similar text contents to the querying sample, where the embeddings of these texts are used to calculate the cosine similarity. Here we assume we have access to an high-quality text embedding model at hand, which can transform a piece of text to a separate vector in the semantic space in which the similarity between two vectors reflect their similarity in contents. We consider three kinds of texts:
\begin{itemize}
\item \noindent \textbf{Retrieving via similar prompts (ST-Q).} We use the prompts in the supporting set as the contents to build the vector database, and use the prompt of the test sample as the input text for retrieving, i.e., comparing the similarity between $\text{prompt}_{test}$ and $\text{prompt}_{i}$.  

\item \noindent \textbf{Retrieving via similar prompts \& draft response (ST-PDR).} In this strategy, since the ground truth answer is not available during inference, we can not retrieve demonstrations with the querying sample's answer. However, note that the LVLM itself can generate a draft response by only generating conditioned onthe prompt or using strategy ST-Q. Thus, we first generate a draft response $\text{response}_{test}^{pred, 1}$ to the test sample $x_{test}$, and then compare the semantic similarity between ($\text{prompt}_{test}$, $\text{response}_{test}^{pred, 1}$) and ($\text{prompt}_{i}$, $\text{response}_{i}$). Note that generating the draft response $\text{response}_{test}^{pred, 1}$ introduces additional latency for the whole system. To ensure small lantency, we ask the model to generate at most 2 tokens. 

\end{itemize}

\noindent \textbf{Retrieving via Similar image-texts (SIT).} Besides retrieving via only images or texts, we can also retrieve the demonstrations via the concatenation of image embeddings and text embeddings. Note that \cite{li2024configure} neglect this group of strategy. Since the CLIP model can generate two vectors for the text and image contents separately, these two vectors will be concatenated. 

Thus, similar to the previous strategies based on text input, we can have the following strategy:
\begin{itemize}
\item \noindent \textbf{Retrieving via similar image and prompts (SIT-IP).} We concatenate the querying image embedding and prompt embedding for retrieval on a vector database, which are constructed by concatenating supporting samples' image embeddings and prompt embeddings.   

\item \noindent \textbf{Retrieving via similar image prompt  and draft response (SIT-IPDR).} In this strategy, we first generate a draft response $\text{response}_{test}^{pred, 1}$ to the test sample $x_{test}$ with the help of strategy SIT-IP, and then compare the semantic similarity between ($\text{image}_{test}$, $\text{prompt}_{test}$, $\text{response}_{test}^{pred, 1}$) and ($\text{image}_{i}$, $\text{prompt}_{i}$, $\text{response}_{i}$).

\end{itemize}

In this work, we use the SIT-IPDR strategy as the default strategy for retrieving demonstrations. And we will use experiments (Section \ref{subsec:further_experiments}) to validate this choice.

\subsection{Retriever fine-tuning via LVLM feedback}

The previous sub-section assumes that an embedding model $\mathcal{E}$ is ready to use for any given VL task which can transform the image and text inputs to vectors. Intuitively, one can directly utilize the pre-trained CLIP models \cite{radford2021learning} to initialize $\mathcal{E}$ and obtain the test sample's image or text embeddings, and conduct search for similar demonstrations based on these embeddings. However, as pointed out by the previous works \cite{Li2023UnifiedDR,rubin2021learning}, demonstrations retrieved by an pre-trained embedding model may not benefit the most for the LVLM. Thus, it is natural to consider fine-tuning the embedding model $\mathcal{E}$ so that its retrieved demonstrations better fit the LVLM and help to elicit better responses from the LVLM. 

To train the demonstration retriever, \citet{rubin2021learning} design task-specific training signals for several tasks separately, which makes their methods hard to transfer and scale on various tasks, and hinders systematic and compatible research on demonstration retrieval. For DRUM’s training, we propose to cast the retriever's training signals into
a list-wise ranking loss based on the LVLM's feedback. Then we introduce a training framework in which the retriever iteratively mines high-quality demonstration candidates with the help of the LVLM and learn to rank them in turn. The whole workflow are shown in Algorithm \ref{alg:DRUM}. And we now introduce the list-wise ranking training and iterative mining strategy for the demonstration retrievers as follows. 

For the training process of DRUM, we splits the dataset for the current visual-language task into four parts: the support set $\mathcal{D}_{supp}$, the training set $\mathcal{D}_{clip\_train}$ used for fine-tuning the image-text embedding model, the validation set $\mathcal{D}_{clip\_dev}$ used to validate the image-text embedding model after fine-tuning, and the test set $\mathcal{D}_{test}$ for evaluating the performance of LVLM contextual learning.

\subsubsection{Ranking demonstration candidates by LVLM feedbacks} \quad Given a querying example $x_{q} = $ ($\text{image}_{q}$, $\text{prompt}_{q}$, $\text{response}_{q}$) from $\mathcal{D}_{clip\_train}$ and its candidate demonstrations $\{ z_i \}_{i=1}^{n}$, we rank these candidates by:
\begin{equation}
\begin{aligned}
r(z_j) & = \text{Ranking}(s(z_j) | \{ s(z_j) \}_{i=1}^{n} ) \\
s(z_j) & = \text{LLH}(\text{response}_{q} |z_j, \text{image}_{q}, \text{prompt}_{q} ), \\  
\end{aligned}
\label{eq:demons_llm_ranking}
\end{equation}
where $\text{LLH}(\cdot | \cdot)$ is the LVLM's conditional log-likelihood function, and $\text{Ranking}$ is the ranking function (in ascending order). We first calculate $s(z_j)$ as the ground-truth $\text{response}_{q}$'s log-likelihood conditioned on the demonstration candidate $z_j$ and the querying input $\text{image}_{q}$ and $\text{prompt}_{q}$. $s(z_j)$ indicates the importance of $z_j$ for the LVLM to encode the querying sample and generate the ground-truth response. Then we rank the demonstration candidates according to $\{ s(z_j) \}_{i=1}^{n}$. The more important $z_j$ is for $x$, the higher $z_j$'s rank $r(z_j)$ will be, or the smaller $r(z_j)$ will be. 

Note that in real-world online applications, one can not re-rank the retrieved candidates according to Eq \ref{eq:demons_llm_ranking}, since it will introduce unbearable latency to users. Thus, in order to maintain the system's efficiency while improving the demonstration sequence's quality, our next step will be to fine-tune the demonstration retriever, so that its ranking of the retrieved demonstrations will be more consistent with that of the LVLM's feedback.  


\subsubsection{Loss function for the demonstration retriever} \quad Note that upon retrieval with the embedding model $\mathcal{E}$, the demonstration candidates of the querying sample $x_{q}$ already have a ranking (in ascending order) which is based on the embeddings' similarity scores, and this ranking is directly reflected by the candidates' indices in $\{ z_j \}_{i=1}^{n}$:
\begin{equation}
r_0(z_j) = \text{Ranking}(j | \{ z_j \}_{i=1}^{n} ),
\label{eq:rank_by_clip}
\end{equation}
and the fine-tuning aims at improving the consistency between $\{ r(z_j) \}_{i=1}^{n}$ and $\{ r_0(z_j) \}_{i=1}^{n}$. Note that we unify different VL tasks’ training signals for the demonstration retriever into the unified list-wise ranking loss using LM’s feedback, instead of designing task-specific objectives like \cite{rubin2021learning}.

With these candidates’ ranks $\{ r(z_j) \}_{i=1}^{n}$ from the LVLM’s feedback, we propose to use the following loss function to inject the ranking signal into the demonstration retriever $\mathcal{E}$:
\begin{equation}
\mathcal{L}_{r} = \sum_{1 \leq i, j \leq n, i \neq j} m(i, j) * \text{log} ( 1 + e^{ (\text{sim}(x_{q}, z_j) - \text{sim}(x_{q}, z_i) ) } ),
\label{eq:loss_function}
\end{equation}
where $m(i, j)$ is given by 
\begin{equation}
m(i, j) = \text{max} (0, \dfrac{1}{ \sqrt{ r(z_j) } } - \dfrac{1}{ \sqrt{ r(z_i) } } ).
\label{eq:weight_coefficient_in_loss}
\end{equation}
For those $z_i$ and $z_j$ where $r(z_j) \leq r(z_i)$, $\mathcal{L}_{r}$ will draw $\text{sim}(x_{q}, z_i)$ up and optimize the retriever towards $\text{sim}(x_{q}, z_i) > \text{sim}(x_{q}, z_j)$. For $z_i$ and $z_j$ where $r(z_i) \geq r(z_j)$, this pair will be discarded by the loss function. Additionally, $m(i, j)$
adjusts the weight for each pair of demonstrations, conveying list-wise ranking information into $\mathcal{L}_{r}$. When the ranks of $z_i$ and $z_j$ are close, e.g., $r(z_i) = 2$ and $r(z_j) = 1$, $m(i, j) \approx 0.292$. In comparison, when $z_i$ has a much higher rank than $z_j$, e.g., $r(z_i) = 15$ and $r(z_j) = 1$, $m(i, j)$ will be $0.742$, larger than 0.292. Thus, when $z_i$ has a much higher rank than $z_j$, $w$ will be a high weight, and asks $\mathcal{L}_{r}$ to strongly draw $\text{sim}(x_q, z_i)$ up and away from $\text{sim}(x_q, z_j)$. Since we optimize the retriever on demonstration pairs under different $m(i, j)$, $\mathcal{L}_{r}$ can help our DRUM method fully incorporate candidates’ list-wise ranking signals and learn to retrieve those high-quality and helpful demonstrations.

\subsection{Iterative Demonstration Candidate Mining}
  
The selection of demonstration candidates can be a key factor for retriever's training. It is infeasible and possibly harmful to take the entire training set as candidates. In addition, once the embedding model is fine-tuned, it no longer matches the supporting samples' vectors in the vector database. To strike a balance between training time cost and quality, we adapt an iterative strategy to update candidates \cite{Li2023UnifiedDR}.
Specifically, we iteratively train the retriever and
use it to select candidates in turn. At each iteration, we update each supporting example $x_q$’s candidates as:
\begin{equation}
Z^{*} = \text{topK} ( \{ \text{sim} (x_q, z) | z \in D \}, n ) ,
\end{equation}
where $D$ is the task’s supporting set, $n$ is the number of candidates retrieved. Then we will use the LVLM $\mathcal{M}$ to score and rerank $Z^{*}$, and calculate the list-wise ranking loss according to Eq \ref{eq:loss_function}. Before the first iteration, the retriever is exactly the pre-trained embedding model, so we initialize candidates based on the similarity calculated with the pretrained embedding model. In summary, Algorithm \ref{alg:DRUM} shows the DRUM’s overall training procedure.

\begin{algorithm}[!ht]
\DontPrintSemicolon

\KwInput{Embedding model $\mathcal{E}$, large vision-language model $\mathcal{M}$, number of training iterations $N_1$, number of training steps in each iteration $N_2$, number of retrieved candidates $n$}
\KwOutput{A fine-tuned embedding model $\mathcal{E}$. }
\KwData{support set $\mathcal{D}_{supp}$, model $\mathcal{E}$'s training set $\mathcal{D}_{clip\_train}$, model $\mathcal{E}$'s validation set $\mathcal{D}_{clip\_dev}$, test set for the LVLM $\mathcal{D}_{test}$; }

training iteration index $i \leftarrow 0$;

\While{$i < N_1$} 
{
    Embed each training example with $\mathcal{E}$;
    
    Retrieve $n$ candidates of each training example;
    
    training step index $j \leftarrow 0$;
    
    \While{$j < N_2$} 
    {

        Sample an querying example $x_q$ from $\mathcal{D}$, and obtain its candidates $\{ z_k \}_{k=1}^{n}$;
        
        Re-rank $\{ z_k \}_{k=1}^{n}$ by $\mathcal{M}$ using Eq \ref{eq:demons_llm_ranking};
        
        Calculate $\mathcal{L}_{r}$ using Eq \ref{eq:loss_function};
        
        Update $\mathcal{E}$;

        $j \leftarrow j + 1$;
    }

    $i \leftarrow i + 1$;

}

\caption{DRUM's demonstration ranking training}
\label{alg:DRUM}

\end{algorithm}

\begin{equation}
S(\mathcal{E}) =  \dfrac{ \sum_{ x_q \in \mathcal{D}_{clip\_dev} } \text{corr}_q }{  \| \mathcal{D}_{clip\_dev} \|  }. 
\end{equation}

\noindent \textbf{Embedding Model Validation} \quad The optimization objective of model $\mathcal{E}$ is to minimize the discrepancy between the ranking of retrieved example vectors and the ranking assigned by the large-scale model $\mathcal{M}$ to these examples. Therefore, to validate the training effectiveness of model $\mathcal{E}$, this work computes the average correlation coefficient of rankings using dataset $\mathcal{D}_{clip\_dev}$. For each sample $x_q$ in $\mathcal{D}_{clip\_dev}$, examples are retrieved using the current model $\mathcal{E}$ and then reranked using the large-scale model $\mathcal{M}$, resulting in two rankings $\{ r_{0}(z_j) \}_{i=1}^{n}$ (Eq~\ref{eq:rank_by_clip}) and $\{ r(z_j) \}_{i=1}^{n}$ (Equation~\ref{eq:demons_llm_ranking}). The correlation coefficient between these two rankings is denoted as $\text{corr}_q$. Thus, the evaluation metric for model $\mathcal{E}$ is:
\begin{equation}
S(\mathcal{E}) =  \dfrac{ \sum_{ x_q \in \mathcal{D}_{clip\_dev} } \text{corr}_q }{  \| \mathcal{D}_{clip\_dev} \|  }. 
\label{eq:embedding_model_eval}
\end{equation}

\section{Experiments}

\subsection{Datasets} 

We conduct experiments on three benchmark visual question-answering (VQA) tasks, two image classification (ImageCLS) tasks, and two image captioning (ImageCAP) tasks: VQAv2 \cite{goyal2017making}, VizWiz \cite{gurari2018vizwiz}, OK-VQA \cite{marino2019ok}, Flowers102 \cite{nilsback2008automated}, Hateful-Memes \cite{kiela2020hateful}, Flickr30K \cite{plummer2015flickr30k}, NoCaps \cite{agrawal2019nocaps}. The introduction and dataset splits of each dataset are detailed in the Supplementary Material.

\subsection{Evaluation metrics}

\noindent\textbf{Metric for the VQA tasks} \quad We follow \citet{alayrac2022flamingo} to use accuracy as the evaluation metric for VQA task. The detailed calculation formula is as follows:
\begin{equation}
\text{Acc}_{a_{i}} = \text{min} (1, \dfrac{ 3 \times \sum_{k \in [0, 9]} \text{match}(a_i, g_k) }{10}),
\end{equation}
where $a_{i}$ denotes the predicted answer of the LVLM, $g_k$ denotes the $k$-th ground true answer, and the $\text{match}()$ function indicates whether two answers match, if they match, the result is 1, otherwise it is 0.

\noindent\textbf{Metric for the image classification tasks} \quad For the visual classification tasks, we report the accuracy score.

\noindent\textbf{Metric for the image captioning tasks} \quad For evaluation on the image captioning tasks, we report the ROUGE-L score \cite{lin2004rouge}.

\begin{table*}[tb!]
\centering
\resizebox{0.78\textwidth}{!}{
\renewcommand\arraystretch{1.2}
\begin{tabular}{c|ccccccc}

\hline
\textbf{Retrieval}   &   \multicolumn{3}{c}{\textbf{VQA}}     &     \multicolumn{2}{c}{\textbf{ImageCLS}}     &     \multicolumn{2}{c}{\textbf{ImageCap}}  \\ 

\textbf{Methods} &  VQAv2   &     VizWiz    &    OK-VQA    &    Flowers102   &    Hateful-Memes     &  Flicker30K   &   NoCaps     \\
\hline

Null   &   36.7    &    14.3    &   11.8     &    13.4   &  44.6   &    17.3    &    19.6  \\

Random   &   46.2  &     33.6    &   26.3     &   31.3    &    51.3     &    27.5   &  29.8    \\
Fixed    &    46.3    &    32.4    &  27.8    &    32.0   &  51.1    &   28.7    &  29.6     \\

BM25    &   48.4    &   24.8    &   25.2    &      25.6  &   46.5    &   23.8    &  24.6    \\

Dino     &   49.3   &    36.8    &  29.9    &       35.7  &    53.2    &   29.0    &   28.8  \\
BGE      &   48.7   &    27.9     &    31.6    &    26.3 &  46.7  &     23.7     &  24.8   \\

CLIP    &   49.9    &   48.2   &   33.4     &    36.5   &  55.4   &   29.2   &  30.7   \\

EPR   &   50.6   &    51.2   &    34.7    &     38.7  &       
 56.7    &  30.1    &   31.6   \\


\hline

Dr-VL   &   \textbf{52.4}   &   \textbf{54.6}    &   \textbf{36.8}     &    \textbf{40.2}   &   \textbf{59.6}    &    \textbf{31.7}     &   \textbf{33.9}   \\

\hline

\end{tabular}}

\caption{\label{tab:results_main_1} Results on 7 benchmark tasks. Due to randomness, the results from Random, Fixed, EPR, UDR and DRUM are the average scores across five different runs under different random seeds. Best scores are bolded. } 

\end{table*}

\subsection{Implementation details}

\noindent \textbf{Computing infrastures} \quad All experiments are conducted on the RTX 4090 GPUs with FP16 precision. 

\noindent \textbf{LVLM models} \quad We employ the Open-Flamingo \cite{lin2004rouge} model (9B) as the LVLM to evaluate our DRUM method.

\noindent\textbf{Decoding} \quad After receiving the input images and text prompts, the predictions are generated using the language modeling head (LM head) of the LVLM. No other prediction layers outputting numerical or categorical results are installed on the LVLM backbone. For decoding during inference, we use beam search with beam size 3.

\textbf{ICL Setup for the LVLM Model $\mathcal{M}$} \quad The number of demonstrations obtained for each test sample is set by default to $n = 4$ in this work. The ablation studies also investigate different values of $n$. After retrieving the examples, model $\mathcal{M}$ concatenates the demonstration sequence in ascending order of similarity scores to the left side of the test sample input. This means that the higher the similarity score an retrieved example has, the closer it is placed to the test sample input.

\noindent\textbf{Settings for embedding and retrieval} \quad This work defaults to using the base-sized CLIP model\footnote{\url{https://huggingface.co/openai/clip-vit-base-patch32}} for image-text embedding. The default retrieval strategy adopted in this work is the SIT-IPDR approach detailed in Section~\ref{subsec:demon_retrievals}. Under this strategy, the vector representation of both demonstrations samples and test samples is obtained by concatenating the image vector and the text vector. This work utilizes the Faiss toolkit~\cite{douze2024faiss} for constructing the vector database and for efficient vector retrieval.

\noindent\textbf{Settings for fine-tuning CLIP model} \quad We set the number of training iterations $N_1$ to 50, the number of training steps in each iteration to $100$. The optimizer is AdamW \cite{Loshchilov2019DecoupledWD}, the learning rate is 1e-5, and the number of warmup steps is set to 50.

\noindent\textbf{Settings for fine-tuning the embedding model} \quad We implements the fine-tuning process of the embedding model $\mathcal{E}$ based on the Huggingface Transformers\cite{wolf-etal-2020-transformers} code library. The number of training epochs $N_1$ for the embedding model is set to 50, with $N_2 = 100$ steps per epoch. During the fine-tuning of the embedding model, the number of recall examples $n$ is set to 32. For model optimization, we use AdamW \cite{Loshchilov2019DecoupledWD}, with a learning rate of 1e-5 and a warmup of 50 steps at the beginning of the model fine-tuning. Other hyperparameters remain consistent with the Transformers code library. After each epoch, the embedding model $\mathcal{E}$ is evaluated according to Equation~\ref{eq:embedding_model_eval}. The fine-tuning employs an early stopping strategy with a maximum patience of 10, meaning that if the evaluation metric $\mathcal{S}(\mathcal{E})$ does not improve for 10 consecutive epochs, the training will be stopped.

\subsection{Baseline methods} 

With the same inference LVLM, Open-Flamingo 9B \cite{awadalla2023openflamingo}, we compare our DRUM method with existing methods for demonstration retrieval by the downstream ICL performance, including: (a) Null, which is not to use any demonstrations. (b) Random, randomly sampling demonstrations from the supporting set. (c) BM25, a prevailing sparse retriever widely used in the literature \cite{chen2017reading}. (d) DINO, which is to retrieve demonstrations using the image embedding provided by the DINO model \cite{caron2021emerging}. (e) BGE, which is to retrieve demonstrations using the text embedding provided by the BGE model \cite{chen2024m3}. (f) CLIP, which is to retrieve demonstrations using the image-text embedding provided by the CLIP model \cite{caron2021emerging}. (g) EPR \cite{rubin2021learning}, which builds upon the aforementioned CLIP approach by conducting LVLM feedback evaluation for each example, then transforming the task of re-ranking demonstrations into a classification task, leading to the training of a classifier for evaluating these demonstrations.

\subsection{Main Results}

We report the performance of different methods on the seven benchmark VL tasks in Table \ref{tab:results_main_1}. We can see that: (a) DRUM outperforms the baselines with clear margins on most tasks, which shows our method’s best demonstration retrieval ability on a wide range of VL tasks. (b) Specially, compared with EPR, DRUM has better overall performance and this shows the effectiveness of our training method. Meanwhile, compared with CLIP, the embedding model which is directly initialized with CLIP-base, DRUM has clear advantages. This straightly demonstrates that our proposed training framework can help DRUM incorporate LVLM’s feedback through the DRUM's fine-tuning procedure and retrieve more beneficial demonstrations. The experimental results also reveal that the random baseline achieves the worst performance in most tasks. This phenomenon is intuitive: pairing the current query with irrelevant demonstrations is unhelpful, and sometimes could lead the model to the wrong directions.

\subsection{Further analysis} 
\label{subsec:further_experiments}

\noindent \textbf{Ablation Study} \quad To evaluate the effect of our DRUM's each component, we consider the following variant of DRUM: (a) DRUM-1, which substitute Eq~\ref{eq:weight_coefficient_in_loss} to $m(i, j) = \text{max}(0,  \dfrac{1}{r(z_j)} - \dfrac{1}{r(z_i)} )$. (b) DRUM-2, which substitute Eq~\ref{eq:weight_coefficient_in_loss} to $m(i, j) = \text{max}(0,  r(z_i) - r(z_j) )$. (c) DRUM-3 removes the weight $m(i, j)$ from Eq~\ref{eq:loss_function}. (d) DRUM-4, which do not conduct iterative demonstration candidate mining. The results are reported in Table \ref{tab:results_ablation}.

The experimental results show that: (a) The comparison between DRUM-1 and DRUM demonstrates the

\begin{table}[tb!]
\centering
\resizebox{0.40\textwidth}{!}{
\renewcommand\arraystretch{1.05}
\begin{tabular}{c|ccc}
\hline
\textbf{Method}   &    \textbf{VizWiz}     &    \textbf{Hateful-Memes}    &    \textbf{Flicker30K}   \\
\hline
DRUM    &   \textbf{54.6}     &   \textbf{59.6}   &   \textbf{31.7}   \\

DRUM-1   &   54.1  &  58.9  &   30.8    \\

DRUM-2   &   54.2  &  59.0  &   30.6     \\

DRUM-3   &   53.7  &   58.6  &   30.3   \\

DRUM-4   &   53.1   &   57.9   &  29.8     \\

\hline
\end{tabular}}
\caption{\label{tab:results_ablation} Results of the ablation study on DRUM's training strategy. }
\end{table}

\begin{table}[tb!]
\centering
\resizebox{0.44\textwidth}{!}{
\renewcommand\arraystretch{1.05}
\begin{tabular}{c|ccc}
\hline
\textbf{Strategy}   &    \textbf{VizWiz}     &    \textbf{Hateful-Memes}    &    \textbf{Flicker30K}   \\ 
\hline
SIT-IPDR    &   \textbf{54.6}     &   \textbf{59.6}   &   \textbf{31.7}   \\
\hdashline

SIT-IP   &    53.1   &  58.3   &   30.8   \\

ST-PDR   &    51.4    &  55.9   &  29.3     \\

ST-P  &   32.8   &  47.0   &   23.8   \\

SI   &   43.6  &   58.3   &  31.1    \\

\hline
\end{tabular}}
\caption{Results of the ablation study on the demonstration retrieval strategy. }
\label{tab:results_ablation_retrieval_strategy} 
\end{table}

\begin{table*}[tb!]
\centering
\resizebox{0.54\textwidth}{!}{
\renewcommand\arraystretch{1.05}
\begin{tabular}{c|c|ccc}
\hline
\textbf{LVLM $\mathcal{M}$}  &  \textbf{$\mathcal{E}$}   &     \textbf{VizWiz}     &    \textbf{Hateful-Memes}    &    \textbf{Flicker30K}   \\ 
\hline

\multirow{3}*{GPT-4o}    &   CLIP   &    65.3     &   66.7  
 &   37.5     \\

   &   CLIP + EPR    &     65.9      &  69.2    &   39.8     \\

  &   CLIP + DRUM    &   \textbf{67.1}     &   \textbf{70.4}    &   \textbf{41.4}    \\

\hline

\multirow{3}*{Claude 3 Opus}    &   CLIP   &     65.5  &   66.3   &   36.9     \\

   &   CLIP + EPR    &    65.6   &  68.8    &   39.7    \\

  &   CLIP + DRUM    &    \textbf{66.8}   &   \textbf{69.9}   &   \textbf{41.6}     \\

\hline
\end{tabular}}
\caption{ Experiments on the transfer learning capabilities of DRUM. We using the fine-tuned model $\mathcal{E}$ to retrieve demonstrations for GPT-4o and Claude 3 Opus. $\mathcal{E}$ being CLIP means no fine-tuning is conducted. $\mathcal{E}$ being CLIP + EPR means fine-tuning with the EPR method is conducted. $\mathcal{E}$ being CLIP + DRUM means fine-tuning with the DRUM method is conducted. } 
\label{tab:results_lvlm_transfer} 
\end{table*}

\begin{figure*}
\centering
\includegraphics[width=0.60\textwidth]{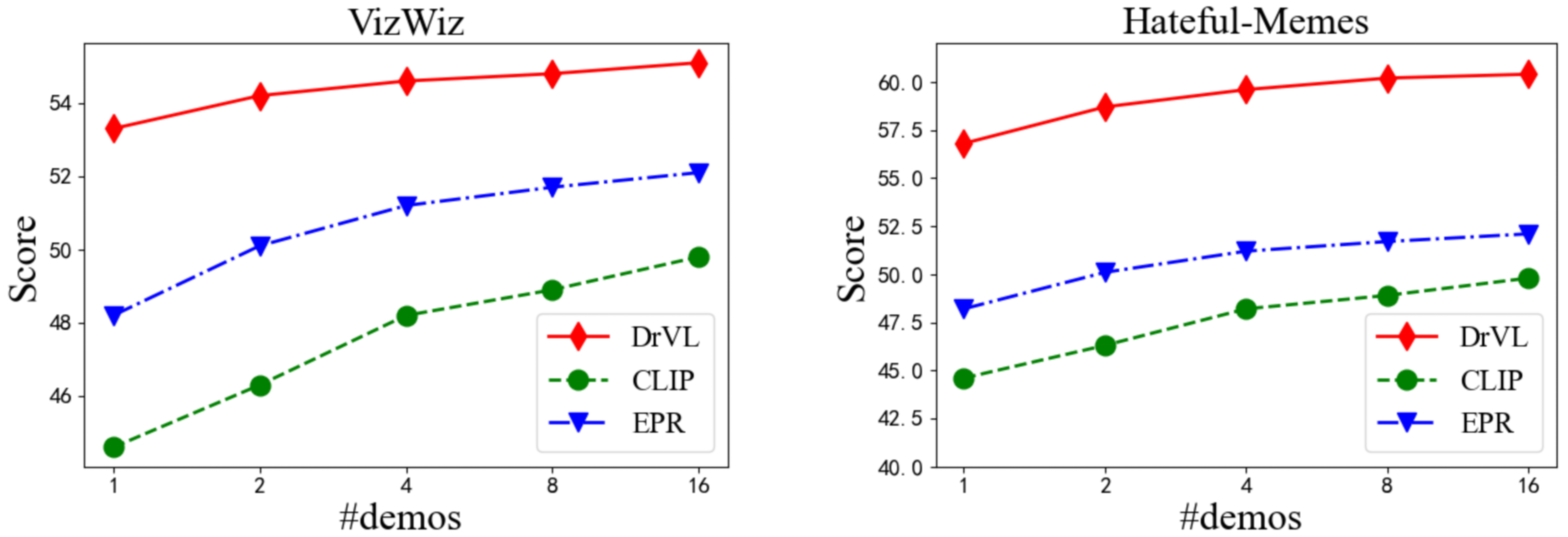}
\caption{The effects of the number of demonstrations on DRUM, EPR, and CLIP. }
\label{fig:DRUM_num_demos} 
\end{figure*}

\noindent \textbf{Ablation on the retrieval strategy} \quad This work uses the SIT-IPDR strategy for example retrieval in the main experiment (Table~\ref{tab:results_main_1}). To demonstrate the rationality of the DRUM setup and this strategy, we conduct an ablation study on the demonstration retrieval strategy. Table~\ref{tab:results_ablation_retrieval_strategy} reports the performance of the DRUM method using SIT-IP, ST-PDR, ST-P, and SI strategies. The experimental results show: (a) The SIT-IPDR strategy outperforms other strategies. This strategy combines image and text information for demonstration retrieval, utilizing the maximum amount of semantic information available in the test sample, thus enabling it to recall the most relevant demonstrations. (b) Retrieving examples based only on the prompt text content (ST-P) performs poorly on image classification tasks and image caption generation tasks. The primary reason for this phenomenon is that these types of tasks involve prompts that contain generic task instructions without directly related semantic information. However, by combining the prompt text with the draft response text (ST-PDR), there is a significant improvement in performance. This result shows that the draft response can effectively supplement the semantic information needed for example retrieval.

\noindent \textbf{Transferability across Different LMs} \quad Note that during the fine-tuning of the embedding model $\mathcal{E}$ using the DRUM method, the LVLM model $\mathcal{M}$ needs to re-rank recalled examples based on conditional likelihood function values. However, if commercial LVLMs such as GPT-4o\footnote{\url{https://openai.com/index/hello-gpt-4o/}}, Claude 3 Opus\footnote{\url{https://www.anthropic.com/news/claude-3-family}}, which are accessed via API, are used in the application, these models do not provide an interface for computing conditional likelihood functions, and thus cannot be used for optimizing the example recall model with the DRUM method. Given that different LVLM models have similar training mechanisms and are pre-trained on large amounts of internet data, their internal mechanisms and cognition share similarities. In this part of the experiment, we will use the embedding model $\mathcal{E}$, fine-tuned with feedback from the Open-Flamingo 9B model, for example recall with GPT-4o or Claude 3 Opus models. The experimental results are presented in Table~\ref{tab:results_lvlm_transfer}.

According to Table \ref{tab:results_lvlm_transfer}, the embedding model, fine-tuned with feedback signals from the Open-Flamingo 9B model, is able to recall higher-quality examples, effectively enhancing the performance of powerful commercial LVLM models like GPT-4o or Claude 3 Opus in tasks such as VQA (Visual Question Answering), image classification, and image caption generation. This experiment demonstrates the practical significance of the DRUM method: by fine-tuning an example recall model with feedback from open-source LVLM models, and then applying this example recall model to the contextual learning of commercial LVLM models.

\noindent \textbf{Impact of demonstration quantity} \quad In the main experiments (Table \ref{tab:results_main_1}), we set $n$ to 4. We now compare DRUM with CLIP and EPR under different amounts of demonstrations, and the experimental results are reported in Figure \ref{fig:DRUM_num_demos}.

We can see that DRUM outperforms baselines consistently across varying amounts of demonstrations. Meanwhile, we can draw two conclusions from the results: (a) The number of demonstrations has a greater impact on the generation task, VizWiz, than the classification task, Hateful-Memes. Specifically, as the number of demonstrations increases, VizWiz’ performance gets significant improvements while Hateful-Memes’ has slight improvements. (b) The quality of demonstrations can be more important than their quantity. Specifically, DRUM with one or two demonstrations still outperforms EPR with 4 demonstrations. These observations again reflect the strong demonstration retrieval ability of DRUM.

\section{Conclusion}

In this paper, we propose DRUM, a unifined approach of demonstration retrieval for large vision-language models. To train DRUM, we cast the LVLM's feedback on a demonstration to a unified list-wise ranking formulation, and propose the ranking training framework with an iterative mining strategy to find high-quality candidates. Experiments on three visual question answering tasks, two visual recognition tasks and two image captioning tasks show that UDR significantly outperforms the baseline demonstration retrieval methods. Further analysis show the effectiveness of each proposed components of the DRUM, and the strong transferability of DRUM across different LVLMs (4B to 175B), unseen datasets, and varying demonstration quantities.

\bibliography{custom}
\bibliographystyle{acl_natbib}

\appendix

\end{CJK*}

\end{document}